\documentclass[10pt,twocolumn,letterpaper]{article}

\usepackage[pagenumbers]{wacv} 

\usepackage[dvipsnames]{xcolor}
\usepackage{graphicx}
\usepackage{amsmath}
\usepackage{amssymb}
\usepackage{booktabs}
\usepackage{customcode}
\usepackage{enumitem}
\usepackage{multirow}
\usepackage{subcaption}
\usepackage{arydshln}
\usepackage{colortbl}
\usepackage{algorithm}
\usepackage{algorithmic}

\newcommand{\code}[1]{\textcolor{blue}{\texttt{#1}}}
\usepackage[pagebackref,breaklinks,colorlinks]{hyperref}

\usepackage[capitalize]{cleveref}
\crefname{section}{Sec.}{Secs.}
\Crefname{section}{Section}{Sections}
\Crefname{table}{Table}{Tables}
\crefname{table}{Tab.}{Tabs.}

\def\wacvPaperID{00042} 
\def\confName{WACV}
\def\confYear{2024}

\begin{document}

\title{PETAL\textit{face}: Parameter Efficient Transfer Learning \\ for Low-resolution Face Recognition}

\author{Kartik Narayan\textsuperscript{1}, Nithin Gopalakrishnan Nair\textsuperscript{1}, Jennifer Xu\textsuperscript{2}, Rama Chellappa\textsuperscript{1}, Vishal M. Patel\textsuperscript{1} \and
\textsuperscript{1}Johns Hopkins University, \textsuperscript{2}Systems and Technology Research \and
{\tt \small \{knaraya4, ngopala2, rchella4, vpatel36\}@jhu.edu, jennifer.xu@str.us}\\
{\small \textcolor{magenta}{\url{https://kartik-3004.github.io/PETALface/}}}}

\maketitle

\begin{abstract}
    Pre-training on large-scale datasets and utilizing margin-based loss functions have been highly successful in training models for high-resolution face recognition. However, these models struggle with low-resolution face datasets, in which the faces lack the facial attributes necessary for distinguishing different faces. Full fine-tuning on low-resolution datasets, a naive method for adapting the model, yields inferior performance due to catastrophic forgetting of pre-trained knowledge. Additionally
    the domain difference between high-resolution (HR) gallery images and low-resolution (LR) probe images in low resolution datasets leads to poor convergence for a single model to adapt to both gallery and probe after fine-tuning. 
    To this end, we propose PETAL\textit{face}, a Parameter-Efficient Transfer Learning approach for low-resolution face recognition. Through PETAL\textit{face}, we attempt to solve both the aforementioned problems. (1) We solve catastrophic forgetting by leveraging the power of parameter efficient fine-tuning(PEFT). (2) We introduce two low-rank adaptation modules to the backbone, with weights adjusted based on the input image quality to account for the difference in quality for the gallery and probe images.  
    To the best of our knowledge, PETAL\textit{face} is the first work leveraging the powers of PEFT for low resolution face recognition.   Extensive experiments demonstrate that the proposed method outperforms full fine-tuning on low-resolution datasets while preserving performance on high-resolution and mixed-quality datasets, all while using only 0.48\% of the parameters.
\end{abstract}
\begin{figure}
    \centering
    \includegraphics[width=\linewidth]{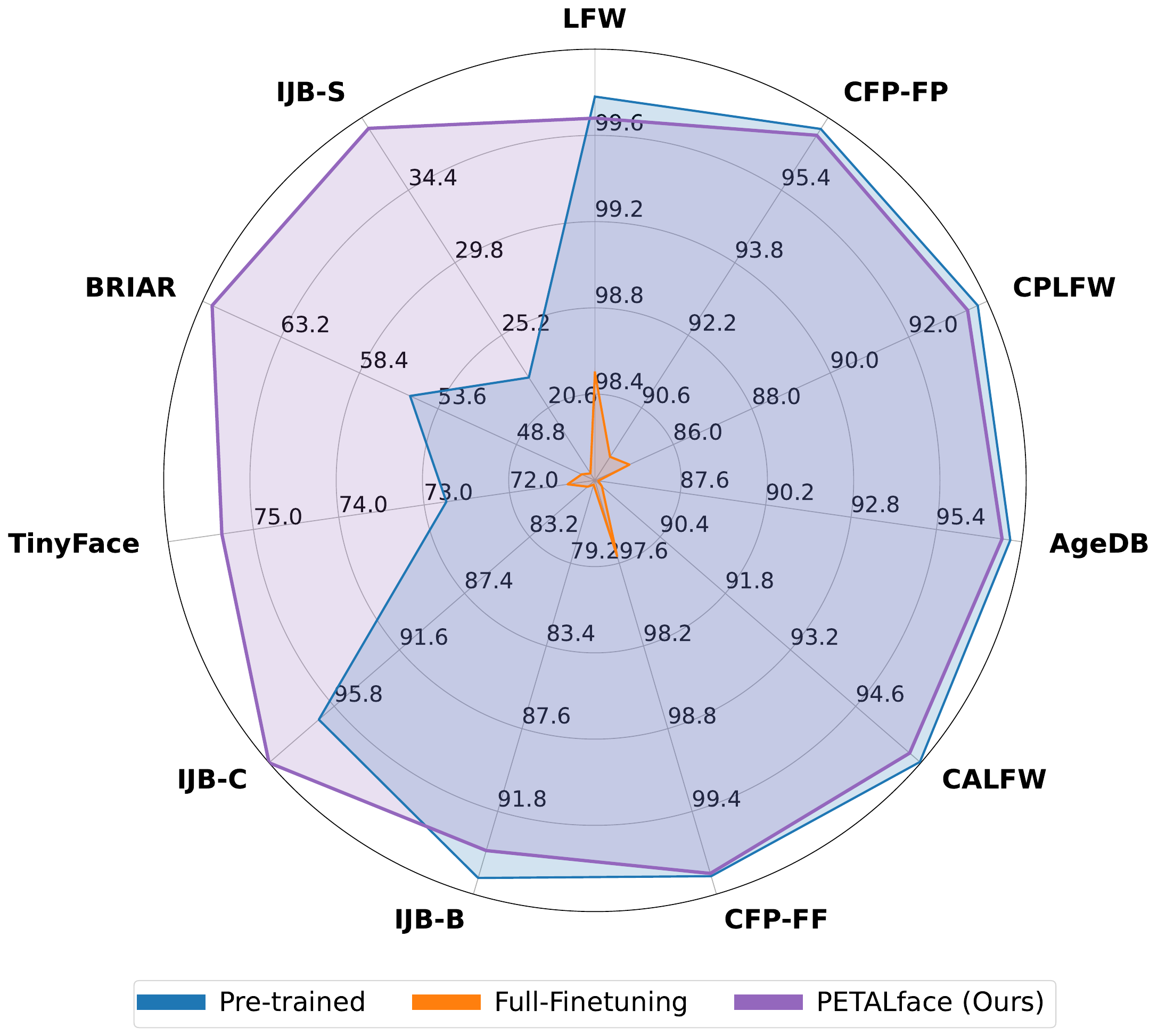}
    \caption{The proposed PETAL\textit{face}: a parameter efficient transfer learning approach adapts to low-resolution datasets beating the performance of pre-trained models with negligible drop in performance on high-resolution and mixed-quality datasets. PETAL\textit{face} enables development of generalized models achieving competitive performance on high-resolution (LFW, CFP-FP, CPLFW, AgeDB, CALFW, CFP-FF) and mixed-quality datasets (IJB-B, IJB-C), with big enhancements on low-quality and surveillance quality datasets (TinyFace, BRIAR, IJB-S). }
    \label{fig:radar}
\end{figure}
\section{Introduction}
\begin{figure*}[t]
    \centering
    \includegraphics[width=\linewidth]{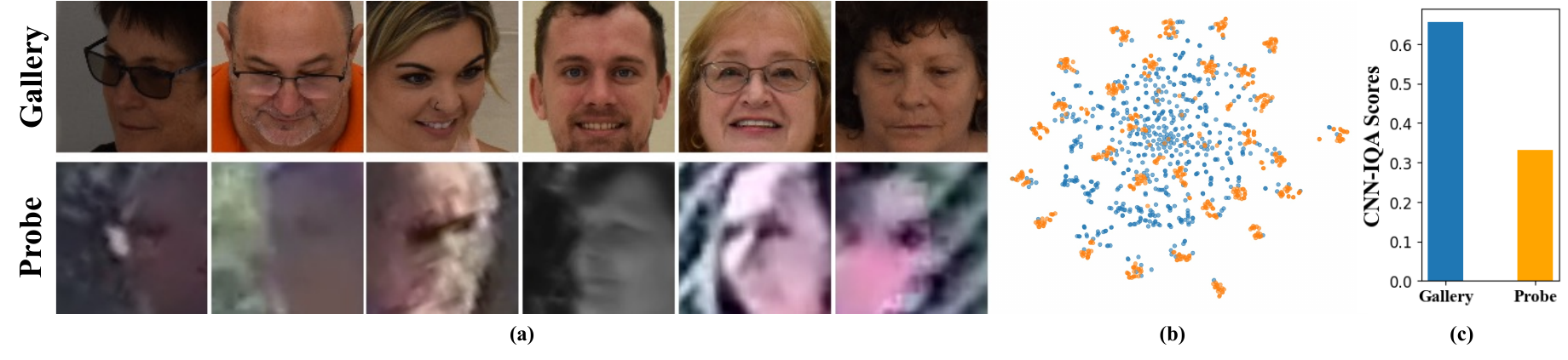}
    \caption{ (a) An illustration of the gallery and probe images from low-resolution dataset (BRIAR). Gallery images usually are high quality compared to the probe images. (b) t-SNE plot for the gallery and probe images of the BRIAR dataset. (c) Average CNN-IQA scores of gallery and probe images for 50 identities of the BRIAR dataset.}
    \label{fig:tsneplot}
\end{figure*}
Face recognition (FR) is one of the primal tasks in biometrics and has been extensively studied for decades due to its importance in device authentication, banking, finance, healthcare, social media, entertainment, retail, marketing, border control, security and surveillance. Early face recognition methods are evaluated using high-quality evaluation datasets and existing state-of-the-art face recognition methods have saturated these benchmarks, with several works achieving over 98\% verification accuracy on high-resolution face recognition datasets like LFW~\cite{huang2008labeled}, CFP-FP~\cite{sengupta2016frontal}, CALFW~\cite{zheng2017cross} and AgeDB~\cite{moschoglou2017agedb}. Recent efforts in face recognition~\cite{cheng2019low, jawade2023conan} have shifted to low-quality face recognition because of its widespread use in surveillance-related applications. Moreover, analysis and generalizability of current methods in low-resolution face-datasets give a measure of the robustness of the recognition algorithm.

Low-resolution datasets~\cite{cheng2019low, kalka2018ijb} contain images with poor clarity as shown in Figure~\ref{fig:tsneplot}(a), making it challenging to extract meaningful discriminative features essential for face recognition and verification. Common degradations include images with low resolution, compression artifacts, motion blur, occlusion, lighting variations, and atmospheric turbulence. Consequently, deep networks trained on high-resolution datasets perform poorly on low-resolution datasets. Moreover, low-resolution datasets are usually small, with a limited number of subjects, as curating them requires significant time, effort, and investment. There are very few low-resolution face recognition datasets that exist, most of which are private. Therefore, low-resolution face recognition remains an unsolved problem with significant room for improvement. Although, there has been a transition to datasets such as IJB-B~\cite{whitelam2017iarpa} and IJB-C~\cite{maze2018iarpa}, and very-low resolution datasets like TinyFace~\cite{cheng2019low}, BRIAR~\cite{cornett2023expanding} and IJB-S~\cite{kalka2018ijb}, research efforts remain focused on margin-based loss functions that create separable identity clusters on the hypersphere. 
 
Existing methods~\cite{kim2022adaface, huang2020curricularface} force the learning of high-resolution and low-resolution face images in a single encoder, failing to account for the domain differences between them, which contradicts our belief. We claim that high-resolution and low-resolution images have distinct distributions and require separate encoders to extract meaningful features for classification. Figure~\ref{fig:tsneplot}(b) shows t-SNE visualization of the BRIAR dataset, where the clusters of gallery (high-resolution) and probe images (low-resolution) images are clearly separated, highlighting the domain difference between them. The bar plot shown in Figure~\ref{fig:tsneplot}(c) further supports this claim showing a clear difference in CNN-IQA scores between the gallery and probe images. This validates our claim that high-quality gallery images and low-quality probe images belong to distinct domains. A straightforward solution is to train two separate encoders for high-resolution and low-resolution data, but this creates misalignment in the embedding space as the two encoders do not share a common final layer. 

A naive approach to adapting pre-trained models to low-resolution datasets is supervised full fine-tuning on these datasets. However, as mentioned, low-resolution datasets are small in size, and updating a model with a large number of parameters on a small low-resolution dataset results in poor convergence. This makes the model prone to catastrophic forgetting and we see a drop in performance on high-resolution and mixed-quality datasets. We illustrate this phenomenon and highlight the drop in performance in Figure \ref{fig:radar}.
Existing methods perform poorly in low-resolution face recognition due to the following issues: 1) small training sets of low-resolution datasets, 2) domain differences between low-resolution and high-resolution data, and 3) catastrophic forgetting while fine-tuning for low-resolution datasets. 

To address the above challenges, we propose a parameter-efficient transfer learning technique called PETAL\textit{face}, which utilizes low-rank adaptation (LoRA)~\cite{hu2021lora} of attention layers to adapt the pretrained model to low-resolution datasets. We introduce two low-rank adaptation modules that are constrained during training and act as separate proxy encoders for high-resolution and low-resolution data, respectively, with a common final embedding layer that helps avoid misalignment in the embedding space. The final output of the model depends on the weightage of these two modules, which is determined based on the image-quality scores of the input images. These scores are provided by an off-the-shelf NR-IQA network and passed to the model along with the images. The use of LoRA ensures that only a small number of parameters are added and trained, drastically reducing the training time. Low-rank adaptation preserves the feature extraction capabilities learned from the pre-training dataset and maintains performance on high-resolution and mixed-quality datasets, resulting in an efficient transfer to low-resolution datasets. The key contributions of our work are summarized below:
\begin{itemize}[noitemsep]
    \item We introduce the use of the LoRA-based PETL technique to adapt large pre-trained face-recognition models to low-resolution datasets. 
    \item We propose an image-quality-based weighting of LoRA modules to create separate proxy encoders for high-resolution and low-resolution data, ensuring effective extraction of embeddings for face recognition. 
    \item We demonstrate the superiority of PETAL\textit{face} in adapting to low-resolution datasets, outperforming other state-of-the-art models on low-resolution benchmarks while maintaining performance on high-resolution and mixed-quality datasets.
\end{itemize}

\section{Related Work}
\textbf{Face Recognition.} Face recognition has significantly progressed from using hand-crafted features~\cite{belhumeur1997eigenfaces, ahonen2006face} to utilizing deep learning models~\cite{Schroff_2015_CVPR, wang2017normface, duan2019uniformface, zhang2019adacos}.
Several works~\cite{deng2019arcface,wang2018cosface,liu2017sphereface, wen2016discriminative} propose different variants of margin-based loss functions for face recognition that show impressive performance on high-resolution benchmarks~\cite{moschoglou2017agedb, huang2008labeled, cfp-paper}. However, much less attention has been given to low-resolution unconstrained face recognition benchmarks~\cite{cheng2019low, kalka2018ijb, cornett2023expanding}, which contain face images that are sometimes unidentifiable due to extreme degradations. To address this, some approaches~\cite{kim2022adaface, huang2020curricularface} incorporate adaptiveness in their training or loss functions to effectively leverage the low-quality images in large datasets~\cite{zhu2021webface260m, guo2016ms}, based on the utility and quality of the low-resolution face images. 

\textbf{Low Resolution Face-Recognition.}
The main challenge in low-resolution face recognition is the domain difference between high-resolution gallery images captured in controlled environments and degraded probe images from surveillance cameras.~\cite{singh2018identity, yue2016image} use super-resolution (SR) models to upscale low-resolution images to high-resolution images to close the domain gap between gallery and probe images. However, several other works~\cite{li2019low, zhang2018super, jiang2018deep} suggest that this approach causes identity hallucination. Many studies~\cite{hsu2019sigan, yin2020fan, yu2018super, singh2021derivenet} have followed, relating recognition to visual quality. However, this is infeasible as it requires paired high-resolution and low-resolution images of the same subject, which are mostly unavailable in low-resolution datasets.~\cite{ massoli2020cross, zhu2019low} use knowledge distillation to transfer knowledge from the high-resolution domain to the low-resolution domain.~\cite{ge2018low} utilizes a teacher-student configuration with help of synthetically degraded samples.~\cite{ge2020look} achieves cross-resolution distillation by employing an additional network between student and teacher network.~\cite{huang2020improving} proposed distribution distillation loss and ~\cite{low2022implicit, low2021mind} introduced augmentations to mitigate the performance gap between high-resolution and low-resolution samples. In our work, we adapt high-resolution model to low-resolution images by parameter efficient transfer learning, employing low-rank adaptation modules weighted based on the image quality of the input. 

\textbf{Parameter-Efficient Transfer Learning.} Parameter-efficient transfer learning was initially introduced in the field of NLP~\cite{houlsby2019parameter, hu2021lora, karimi2021compacter, zaken2021bitfit, lester2021power}. It aims to achieve competitive performance with full fine-tuning by training only a small fraction of the total number of parameters. Recently, it has been adopted in the field of computer vision for various applications~\cite{chen2022adaptformer, chen2205vision, jia2022visual, jie2022convolutional}. VPT~\cite{jia2022visual} appends learnable prompts to frozen transformer layer. Adapter~\cite{houlsby2019parameter} employs feedforward-down and feedforward-up blocks to adapt the pre-trained model. LoRA~\cite{hu2021lora} leverages the low-rank nature of attention weights and performs matrix decomposition for parameter efficiency. Several variants of LoRA have been proposed since then. DyLoRA~\cite{valipour2022dylora} truncates the up-projection and down-projection matrices in the objective, further reducing the number of trainable parameters. ResLoRA~\cite{shi2024reslora} adds a residual path for stable training. DoRA~\cite{liu2024dora} decomposes pre-trained weight into magnitude and direction components, and efficiently updates the direction component. NOAH~\cite{zhang2022neural} and GLoRA~\cite{chavan2023one} introduce Neural Architecture Search (NAS) to combine different methods. SSF~\cite{lian2022scaling} proposes a scale and shift learnable transformation on features of the pre-trained model. FacT~\cite{jie2023fact} uses a tensorization-decomposition framework to break down the weight increments into lightweight factors. In our work, we use LoRA to fine-tune a pre-trained model on low-resolution face images, resulting in improved performance on low-resolution benchmarks while preserving the knowledge of the pre-trained model.

\section{Proposed Work}
\label{sec:proposed}
\begin{figure*}
    \centering
    \includegraphics[width=\linewidth]{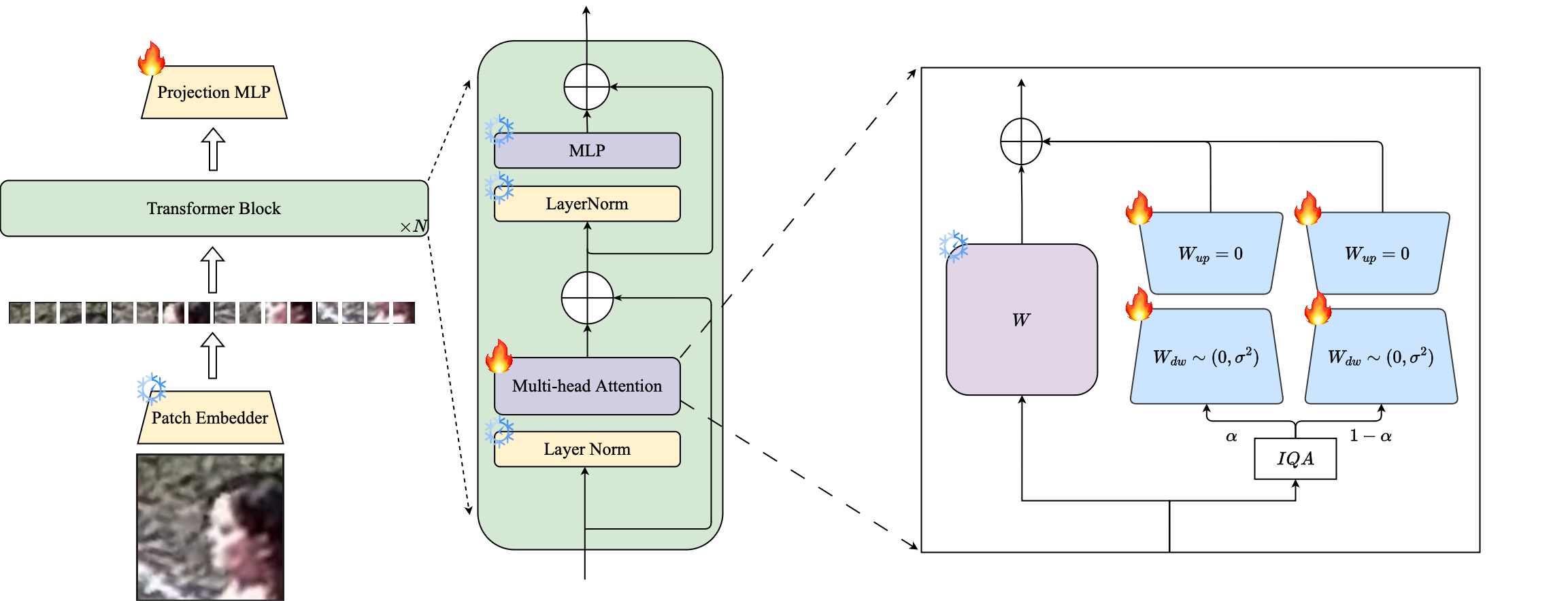}
    \caption{Overview of the proposed \textbf{PETAL\textit{face}}. We include an additional trainable module in linear layers present in attention layers and the final feature projection MLP. The trainable module is highlighted on the right. Specifically, we add two LoRA layers, where the weightage $\alpha$ is decided based on the input-image quality, computed using an off-the-shelf image quality assessment network (IQA).}
    \label{fig:petalface}
\end{figure*}
In this section, we provide the necessary background on the sub-modules utilized in our method, namely  LoRA~\cite{hu2021lora}, followed by a detailed explanation of the proposed fine-tuning procedure: PETAL\textit{face}.

\subsection{Low-Rank Adaptation}
Low-rank adaptation (LoRA)~\cite{hu2021lora}, is a technique that was first introduced to adapt large language modules to low data regime, while retaining the original knowledge learned by a network. To achieve this, additional low-rank parameter modules are added in parallel to the pre-trained weights. During fine-tuning, the original pre-trained weights are kept frozen, and only the LoRA blocks are updated. For a pre-trained weight matrix $W_0 \in \mathbb{R}^{m \times n}$ of a dense layer in the network, LoRA appends a weight update $\Delta W \in \mathbb{R}^{m \times n}$, utilizing a low-rank decomposition such that $\Delta W = W_{up} W_{dw}$, where $W_{up} \in \mathbb{R}^{m \times r}, \, W_{dw} \in \mathbb{R}^{r \times n}, \, \text{and} \, r \ll \min(m, n)$. Here, $r$ is the rank hyper-parameter which controls the bottleneck dimension of the low-rank decomposition. The output $x_{out}$ of the dense layer with input $x_{in}$ can be represented as:
\[
    x_{out} = W_0 x_{in} + \Delta W x_{in} = W_0 x_{in} + \alpha W_{up} W_{dw} x_{in} 
\]
Here $\alpha$ is a constant scale hyper-parameter. In the initial work~\cite{hu2021lora}, $W_{up}$ matrix is initialized with zeros, and the $W_{dw}$ matrix is initialized as a Gaussian distribution with zero mean and standard deviation $1/r$. The zero initialization ensures that during the start of the fine-tuning process, the default configuration corresponds to the pre-trained one. Therefore, any further training should improve the performance over the pre-trained results.

To the best of our knowledge, PETAL\textit{face} is the first to explore parameter-efficient transfer learning methods, such as LoRA, for adapting to low-resolution face recognition datasets. A naive LoRA over a pre-trained transformer network trained for face recognition adds trainable parameters parallel to the attention layers, while the final layer that outputs the embeddings is shared. This shared layer enables the alignment of high-resolution and low-resolution features in the embedding space. It helps alleviate the issue of embedding misalignment that happens while training two separate encoders for gallery and probe. Further, LoRA can be treated as a plug-in module, which could be turned on or off, hence it preserves the pre-trained knowledge of models, performing on par with high-resolution and mixed-quality datasets and on plugging in improves the performance on low-resolution datasets. However, such a naive implementation of LoRA suffers a disadvantage that gallery and probe images share the same parameters even though in most cases gallery images are easier to recognize than probe images. Please refer to Figure \ref{fig:tsneplot}(a). Consequently, if common weights are fit to both datasets, this might lead to an average fit between probe and gallery images. To address this issue of domain difference between gallery and probe images, we propose PETAL\textit{face}, which employs twin LoRA modules weighted based on the input image quality during training. This approach further boosts performance on low-resolution datasets while maintaining performance on high-resolution datasets.

\subsection{PETAL\textit{face}}
PETAL\textit{face} introduces a novel approach for adapting to low-resolution face images using two LoRA (Low-Rank Adaptation) blocks in each adaptation layer of the network as illustrated in Figure~\ref{fig:petalface}. These blocks are constrained during training to ensure that one acts as a proxy encoder for high-resolution images and the other for low-resolution images. We achieve this by assigning different weights to these blocks based on the input image quality, effectively creating proxy encoders tailored to the quality of the input. This dynamic weight assignment allows PETAL\textit{face} to better handle varying input qualities, enhancing overall performance.  For PETAL\textit{face}, we add two LoRA blocks parallel to the attention layer, which are weighted by a parameter in $(0,1)$ depending on the input-image quality. The use of two LoRA blocks, along with the backbone network, enables meaningful extraction of features from both high-resolution and low-resolution images, which is difficult to achieve with a single encoder due to the domain difference, as previously discussed. 
Additionally, we add a LoRA block parallel to the last layer to ensure that the final embeddings are aligned even after adaptation to the domain of the low resolution dataset.

Specifically, let the two LoRA blocks parallel to a pre-trained weight matrix \( W_0 \in \mathbb{R}^{m \times n} \) of a dense layer be \( W_1 \in \mathbb{R}^{m \times n} \) and \( W_2 \in \mathbb{R}^{m \times n} \). Given a batch of \( p \) input images \( X = \{ x_i \mid 0 \leq i < p \} \), PETAL\textit{face} calculates the image quality score using an image quality estimator \(\phi(x)\), represented as $ Q = \{ q_i \mid 0 \leq i < p \ni q_i = \phi(x_i), \forall \ 0 \leq i < p \}$.
For each dataset we fine-tune on, we sample a random \( l \) number of samples \( x_1, x_2, \ldots, x_l \) and calculate an estimate of the mean \( \mu \) and the standard deviation \( \sigma \) for the quality score.
\[
\mu = \frac{1}{l} \sum_{i=1}^{l} \phi(x_i), \quad
\sigma = \sqrt{\frac{1}{l} \sum_{i=1}^{l} (\phi(x_i) - \mu)^2}
\]
We set a threshold \( t = \mu + \sigma \) for the whole dataset and then transform the quality scores \( q_i \) of each sample into weightage \( \alpha_i \) for the LoRA blocks, using the following equation:
\[
\alpha_i = 
\begin{cases} 
0.5 & \text{if } q_i = t \\
0.5 - (t - q_i) & \text{if } q_i < t \\
0.5 + (q_i - t) & \text{if } q_i > t 
\end{cases}
\]
This transformation regularizes the weightages per sample, ensuring that the weightages given to the LoRA blocks are continuous rather than discrete $0$ and $1$. It also stabilizes the training of PETAL\textit{face} and leads to smooth convergence of the loss. The final output is given by:
\[
x_{out} = W_0(x) + \alpha W_1(x) + (1 - \alpha) W_2(x)
\]
Here, based on the weightage $\alpha$,
\[
    \hat{W}(x)=  W_0(x) + \alpha W_1(x) +(1 - \alpha) W_2(x)  
\]
acts as a proxy encoder for high-resolution images as well as low-resolution images. The proposed method allows for separate encoders for different resolutions to exist within a single backbone, differing only by a few low-rank parameters. This approach achieves the dual objectives of resolution-specific encoders and an aligned embedding space, enhancing performance in low-resolution face recognition and preserving pre-trained knowledge.

The low-rank blocks can be added in parallel at various locations. For finding the most suitable layers in a transformer based recognition network to add LoRA blocks, we tested different LoRA placements, as shown in Table~\ref{table:layer_ablation}, and chose the best performing configuration. Specifically, we found that the most effective layers are the attention (qkv) linear weights along with the final feature layer. Additionally, we ablated over various ranks for low-rank decomposition and set it to 8, which delivered superior performance. The rank of a LoRA block is generally set low because the attention matrix has an intrinsic low-rank~\cite{hu2021lora}, which also helps minimize the number of trainable parameters.

\section{Experiments}
\subsection{Datasets}
We employ WebFace4M and WebFace12M~\cite{zhu2021webface260m} as our pre-training datasets, which include about $4$M and $12$M million images, with approximately $205,000$ and $617,000$ distinct identities, respectively. To adapt the model to low-resolution images, we fine-tune it on the training sets of TinyFace~\cite{cheng2019low} and BRIAR~\cite{cornett2023expanding}. We evaluate the fine-tuned models on the test sets of TinyFace~\cite{cheng2019low}, IJB-S~\cite{kalka2018ijb}, and BRIAR~\cite{cornett2023expanding}, demonstrating the superiority of our proposed fine-tuning procedure. TinyFace~\cite{cheng2019low} comprises $169,403$ low-resolution images of $5,139$ identities, with a training subset containing $7,804$ images of $2,570$ identities. IJB-S~\cite{kalka2018ijb} is a surveillance video-based face dataset consisting of $398$ videos and $202$ identities. It is employed under three protocols: \textit{Surveillance-to-Surveillance}, \textit{Surveillance-to-Single}, and \textit{Surveillance-to-Booking}. \textit{Surveillance} refers to surveillance videos, \textit{Single} indicates high-quality enrollment images, and \textit{Booking} includes multiple enrollment images captured from various angles. The BRIAR~\cite{cornett2023expanding} training set consists of $550,000$ images from $577$ unique identities. For evaluation on BRIAR, we adhere to BRIAR Protocol 3.1 (face included treatment)~\cite{cornett2023expanding}. This protocol includes a gallery of $86,958$ controlled images representing $615$ identities, and a probe set comprising $5,435$ clips from $3,441$ unique field videos representing $260$ identities. Additionally, we show that PETAL\textit{face} adapts to low-resolution face images without forgetting the pre-trained knowledge by evaluating it on six high-resolution datasets: LFW~\cite{huang2008labeled}, CFP-FP~\cite{sengupta2016frontal}, CPLFW~\cite{zheng2018cross}, AgeDB~\cite{moschoglou2017agedb}, CALFW~\cite{zheng2017cross}, and CFP-FF~\cite{sengupta2016frontal}, as well as two mixed-quality datasets: IJB-B~\cite{whitelam2017iarpa} and IJBC~\cite{maze2018iarpa}.

\subsection{Evaluation Setup \& Metrics}
\begin{table*}[!t]
\begin{center}
\resizebox{\textwidth}{!}{
\begin{tabular}{@{}lccc cccccc cc ccc@{}}
\toprule[0.15em]
\multirow{3}{*}{\textbf{Training}} & \multirow{3}{*}{\textbf{Loss}} & \multirow{3}{*}{\textbf{Dataset}}& \multirow{3}{*}{\textbf{Arch.}} & \multicolumn{6}{c}{\textbf{High-Resolution}} & \multicolumn{2}{c}{\textbf{Mixed-Quality}} & \multicolumn{3}{c}{\textbf{Low-resolution}} \\
\cmidrule(lr){5-10} \cmidrule(lr){11-12} \cmidrule(lr){13-15}
& &  & & LFW~\cite{huang2008labeled} & CFP-FP~\cite{sengupta2016frontal} & CPLFW~\cite{zheng2018cross} & AgeDB~\cite{moschoglou2017agedb} & CALFW~\cite{zheng2017cross} & CFP-FF~\cite{sengupta2016frontal} & IJB-B~\cite{whitelam2017iarpa} & IJB-C~\cite{maze2018iarpa} & \multicolumn{3}{c}{TinyFace~\cite{cheng2019low}}\\
\cmidrule(lr){5-10} \cmidrule(lr){11-12} \cmidrule(lr){13-15}
& &  & & \multicolumn{6}{c}{Verification Accuracy} & \multicolumn{2}{c}{TAR@FAR=0.01\%} & Rank-1 & Rank-5 & Rank-10 \\
\midrule[0.15em] 
Pre-trained & CosFace~\cite{wang2018cosface} & WBF4M & R50  & 99.68 & 96.83 & 93.28 & 96.88 & 95.63 & 99.70 & 94.09 & 96.01 & 72.71 & 76.36 & 78.99 \\
Pre-trained & ArcFace~\cite{deng2019arcface} & WBF4M & R50 & 99.67 & 96.71 & 93.41 & 96.81 & 95.71 & 99.75 & 94.02 & 95.99 & 73.04 & 76.85 & 79.45 \\
Pre-trained & AdaFace~\cite{kim2022adaface} & WBF4M & R50 & 99.78 & 97.14 & 93.81 & 97.26 & 95.98 & 99.81 & 94.95 & 96.67 & 73.49 & 76.60 & 79.07 \\
\cdashline{1-15}[1pt/1pt]
Pre-trained & CosFace~\cite{wang2018cosface} & WBF4M & ViT-B & 99.73 & 97.30 & 94.31 & 97.51 & 95.95 & 99.87 & 95.18 & 96.87 & 73.57 & 76.95 & 78.94 \\
Pre-trained & ArcFace~\cite{deng2019arcface} & WBF4M & ViT-B & 99.82 & 97.23 & 93.68 & 97.53 & 95.91 & 99.80 & 94.91 & 96.64 & 72.74 & 76.28 & 78.13 \\
Pre-trained & AdaFace~\cite{kim2022adaface} & WBF4M & ViT-B & 99.76 & 97.00 & 93.75 & 96.85 & 95.71 & 99.80 & 94.90 & 96.52 & 74.03 & 77.22 & 79.37 \\
Pre-trained & CosFace~\cite{wang2018cosface} & WBF4M & Swin-B & 99.78 & 96.75 & 93.76 & 97.65 & 95.98 & 99.87 & 95.18 & 96.79 & 72.74 & 76.79 & 79.18 \\
Pre-trained & ArcFace~\cite{deng2019arcface} & WBF4M & Swin-B & 99.76 & 96.77 & 93.93 & 97.35 & 95.83 & 99.87 & 94.87 & 96.66 & 73.31 & 76.68 & 79.23 \\
\midrule[0.15em]
Full-FT & CosFace~\cite{wang2018cosface} & WBF4M & Swin-B & 98.50 & 89.52 & 84.88 & 85.10 & 89.15 & 97.55 & 75.22 & 79.47 & 71.32 & 76.42 & 79.45 \\
Full-FT & ArcFace~\cite{deng2019arcface} & WBF4M & Swin-B & 98.31 & 88.94 & 84.00 & 83.45 & 88.33 & 97.14 & 71.84 & 76.10 & 71.11 & 76.63 & 79.96 \\
LoRA & CosFace~\cite{wang2018cosface} & WBF4M & Swin-B & 99.65 & 96.61 & 93.38 & 97.35 & 95.75 & 99.84 & 93.57 & 95.63 & 75.37 & 78.88 & \textcolor{blue}{\textbf{82.02}} \\
LoRA & ArcFace~\cite{deng2019arcface} & WBF4M & Swin-B & \textcolor{blue}{\textbf{99.73}} & 96.28 & 93.20 & 96.71 & 95.68 & 99.74 & 93.38 & 95.28 & 75.64 & 78.99 & 81.43 \\
\rowcolor[gray]{0.93}
\textbf{PETAL\textit{face}} & CosFace~\cite{wang2018cosface} & WBF4M & Swin-B & 99.68 & \textcolor{blue}{\textbf{96.61}} & \textcolor{blue}{\textbf{93.50}} & \textcolor{blue}{\textbf{97.40}} & \textcolor{blue}{\textbf{95.76}} & \textcolor{blue}{\textbf{99.85}} & \textcolor{blue}{\textbf{93.79}} & \textcolor{blue}{\textbf{95.67}} & 75.45 & \textcolor{blue}{\textbf{79.05}} & 81.19 \\
\rowcolor[gray]{0.93}
\textbf{PETAL\textit{face}} & ArcFace~\cite{deng2019arcface} & WBF4M & Swin-B & 99.66 & 96.37 & 93.18 & 96.45 & 95.61 & 99.80 & 93.29 & 95.27 & \textcolor{blue}{\textbf{75.72}} & 78.86 & 81.70 \\
\midrule
\rowcolor[gray]{0.93}
\textbf{PETAL\textit{face}} & ArcFace~\cite{deng2019arcface} & WBF12M & Swin-B & 99.76 & 97.31 & 94.25 & 98.08 & 95.80 & 99.91 & 95.17 & 96.87 & 76.66 & 79.64 & 81.38 \\
\bottomrule[0.15em]
\end{tabular}
}
\caption{Results of Protocol 1: The models are fine-tuned on train set of TinyFace and tested on several high-resolution, mixed-quality and TinyFace dataset. PETAL\textit{face} adapts to the low-resolution data achieving SOTA results, preserving its performance on other datasets. [\textcolor{blue}{\textbf{BLUE}}] indicates the best results for models trained on WebFace4M~\cite{zhu2021webface260m}.}
\label{table:tinyface}
\end{center}
\vspace{-10pt}
\end{table*}

To validate our proposed claims, we organize our experiments into two protocols. In \textbf{Protocol 1}, we fine-tune our models on the training set of TinyFace and evaluate them on its test set. Additionally, we evaluate the models on high-resolution and mixed-quality datasets. This protocol aims to highlight the capability of PETAL\textit{face} to adapt to low-resolution datasets while maintaining performance on high-resolution and mixed-quality datasets. In \textbf{Protocol 2}, we fine-tune the models on the BRIAR dataset and evaluate them using BRIAR Protocol 3.1 and on IJB-S. We show that PETAL\textit{face} performs better than full fine-tuning and naive LoRA. We evaluate the models on high-resolution and mixed-quality datasets using 1:1 verification accuracy and TAR@FAR at different thresholds, respectively. Rank retrieval (Rank-1, Rank-5, and Rank-10) is used for TinyFace. We report TAR@FAR at different thresholds and closed-set rank retrieval (Rank-1, Rank-5, and Rank-20) for BRIAR. For IJB-S, we report open-set TPIR@FPIR=1\%/10\% and closed-set rank retrieval (Rank-1, Rank-5, and Rank-10).    

\subsection{Implementation Details}
We re-trained all baseline models, consisting of configurations with different backbones (R50, ViT-B, and Swin-N) and loss functions (CosFace, ArcFace, and AdaFace). To ensure a fair comparison, we tested all models on the same cropped and aligned test sets. We fine-tuned the models using the AdamW optimizer with a weight decay of $0.1$. A Polynomial LR scheduler was employed with an initial learning rate of $4e^{-5}$ during training. We sampled $l=1000$ images from the fine-tuning dataset to calculate the mean $\mu$ and variance $\sigma$, which were used to determine the threshold $t$ for calculating weightages for the LoRA blocks.
When fine-tuning on TinyFace, we utilized $2$ warm-up epochs and trained the model for $40$ epochs with a batch size of $8$. For BRIAR, we used $1$ warm-up epoch and fine-tuned for $10$ epochs with the same batch size. We applied a rank of $8$ for low-rank decomposition on TinyFace and $32$ for BRIAR. This difference is due to TinyFace having a relatively smaller training set of approximately $\approx7000$ images, compared to BRIAR's $\approx300$k images. The larger rank for BRIAR increases the number of trainable parameters to accommodate the larger train set.
We employ CNN-IQA~\cite{kang2014convolutional}as our NR-IQA model to classify the images as low-resolution or high-resolution. All code was written in PyTorch, and the models were trained on eight A5000 GPUs, each with $24$GB of memory. The detailed implementation is provided in Appendix~\ref{sec:implementation}.
\section{Results}
In this section, we showcase PETAL\textit{face}'s superiority in transferring to low-resolution datasets maintaining competitive performance on high-resolution and mixed-quality datasets, and compare it with other baselines. We also analyse the benefits of the proposed fine-tuning procedure.
\subsection{Results on Tinyface Dataset-(Protocol-1)}
\label{subsection:tinyface_results}
\begin{table*}[!t]
\begin{center}
\resizebox{\textwidth}{!}{
\begin{tabular}{@{}lcc cccccc ccccc@{}}
\toprule[0.15em]
\multirow{3}{*}{\textbf{Training}}  & \multirow{3}{*}{\textbf{Dataset}}& \multirow{3}{*}{\textbf{Arch.}} & \multicolumn{6}{c}{\textbf{BRIAR Protocol 3.1}~\cite{cornett2023expanding}} & \multicolumn{5}{c}{\textbf{IJB-S (Surveillance to Surveillance)~\cite{kalka2018ijb}}} \\
\cmidrule(lr){4-9} \cmidrule(lr){10-14}
& & & \multicolumn{3}{c}{TAR@FAR} & \multicolumn{3}{c}{Rank Retrieval} & \multicolumn{2}{c}{TPIR@FPIR} & \multicolumn{3}{c}{Rank Retrieval}\\
\cmidrule(lr){4-6} \cmidrule(lr){7-9} \cmidrule(lr){10-11} \cmidrule(lr){12-14}
& & & 0.01\% & 0.1\% & 1\% & Rank-1 & Rank-5 & Rank-20 & 1\% & 10\% & Rank-1 & Rank-5 & Rank-10\\
\midrule[0.15em] 
Pre-trained & WBF4M~\cite{zhu2021webface260m} & R50  & 22.55 & 35.43 & 52.20 & 45.43 & 54.54 & 65.13 & 3.67 & 9.09 & 33.62 & 49.40 & 54.92 \\
Pre-trained & WBF4M~\cite{zhu2021webface260m} & ViT-B  & 34.29 & 47.41 & 62.81 & 55.44 & 64.32 & 73.46 & 2.58 & 8.12 & 25.76 & 40.69 & 47.15 \\
Pre-trained & WBF4M~\cite{zhu2021webface260m} & Swin-B  & 33.77 & 45.93 & 61.17 & 55.31 & 63.29 & 72.76 & 2.11 & 7.45 & 22.52 & 37.97 & 44.93 \\
\midrule[0.15em]
Full-FT & WBF4M~\cite{zhu2021webface260m} & Swin-B  & 11.62 & 29.68 & 58.66 & 44.81 & 59.88 & 74.73 & 1.72 & 5.95 & 16.44 & 31.58 & 38.65 \\
\rowcolor[gray]{0.93}
\textbf{PETAL\textit{face}} & WBF4M~\cite{zhu2021webface260m} & Swin-B  & \textcolor{blue}{\textbf{35.12}} & \textcolor{blue}{\textbf{55.35}} & \textcolor{blue}{\textbf{75.43}} & \textcolor{blue}{\textbf{67.42}} & \textcolor{blue}{\textbf{76.74}} & \textcolor{blue}{\textbf{85.20}} & \textcolor{blue}{\textbf{12.25}} & \textcolor{blue}{\textbf{25.28}} & \textcolor{blue}{\textbf{38.32}} & \textcolor{blue}{\textbf{51.50}} & \textcolor{blue}{\textbf{57.05}} \\
\midrule
\rowcolor[gray]{0.93}
\textbf{PETAL\textit{face}} & WBF12M~\cite{zhu2021webface260m} & Swin-B  & 44.29 & 63.01 & 81.86 & 74.49 & 82.82 & 90.12 & 15.28 & 30.40 & 42.30 & 54.33 & 58.31 \\
\bottomrule[0.15em]
\end{tabular}
}
\caption{Results of Protocol 2: The models are fine-tuned on the BRIAR dataset and tested using BRIAR Protocol 3.1 and the IJB-S dataset. [\textcolor{blue}{\textbf{BLUE}}] indicates the best results for models trained on WebFace4M~\cite{zhu2021webface260m}.}
\label{table:briar}
\end{center}
\vspace{-10pt}
\end{table*}

The results for Protocol-1 are summarized in Table~\ref{table:tinyface}. From the pre-trained models, we observe that different loss functions and backbone architectures result in only minor differences in final performance. We choose the Swin-B~\cite{liu2021swin} architecture for our experiments due to its ability to adapt to out-of-domain distributions~\cite{kim2022broad}. We select ArcFace~\cite{deng2019arcface} for all our experiments as it shows better performance when coupled with Swin-B. Full fine-tuning of pre-trained face recognition models does not lead to performance improvement; instead, we observe a performance decrease from 73.31 to 71.11. Additionally, the performance on high-resolution and mixed-quality datasets also dropped after adaptaion to the low resolution dataset, as can be seen from Table \ref{table:tinyface}. When a model is fully fine-tuned for low-resolution face recognition, it is typically pre-trained on large datasets with millions of identities and then updated based on low-resolution datasets with only a few hundred identities. Due to the domain differences between low-resolution and high-resolution images, the model encounters large gradient updates initially, deviating from the original pre-trained weights suitable for recognition over a large collection of images. This can lead to poor convergence as can be seen from the full fine-tuning results in Table\ref{table:tinyface}. These large gradient updates result in catastrophic forgetting of the pre-trained knowledge, explaining the performance drop for high-resolution and mixed-quality datasets. To further validate these findings, we present a comparison of initial gradients during full fine-tuning and when using the PETAL\textit{face} fine-tuning approach in Appendix~\ref{sec:grad}.

PETAL\textit{face} addresses the problem of catastrophic forgetting and achieves rank-retrieval accuracies of $75.72$\%, $78.86$\%, and $81.70$\% for rank-1, rank-5, and rank-10, respectively. It significantly boosts the performance of pre-trained models while maintaining performance on high-resolution and mixed-quality datasets. We attribute this improvement to the twin low-rank modules added parallel to the attention weights, which are weighted adaptively based on input image quality and extract meaningful features based on the quality of image. The two low-rank modules serve as proxy encoders for high-resolution and low-resolution data, respectively. The adaptive LoRA modules perform better than static LoRA modules, with a performance increase from $75.64$\% to $75.72$\%. Furthermore, as we fine-tune the model by adding weights parallel to the model, they still share the common final embedding layer. This ensures that the feature space for high-resolution and low-resolution data is aligned. As shown in Table~\ref{table:layer_ablation}, we gain a further boost in performance by adding a LoRA module parallel to the final projection MLP. The LoRA module in the projection MLP layer ensures that the embedding space stays aligned by adjusting according to the weight updates of the LoRA modules parallel to the attention layer. Additionally, the low-rank decomposition keeps the trainable parameters to a minimum and makes the fine-tuning process efficient. Finally, we observe that the proposed approach provides better performance metrics when the pre-training dataset is scaled from WebFace4M to WebFace12M, with rank-1 accuracy increasing from 75.72\% to 76.66\%.

\subsection{Results on BRIAR and IJB-S datasets-(Protocol-2)}

With this protocol, we aim to highlight the effectiveness of the proposed PETAL\textit{face} on datasets that have a clear domain difference between gallery (high-resolution) and probe (low-resolution) images. The samples in the TinyFace dataset have similar distributions of gallery and probe images, with a mean CNN-IQA~\cite{kang2014convolutional} score of $60.26$ and a standard deviation of approximately $15.44$. In contrast, the BRIAR and IJB-S datasets have samples with CNN-IQA scores ranging from $20$ to $90$. This highlights that IJB-S and BRIAR are more challenging datasets, demanding a better feature extractor. The results of this protocol are summarized in Table~\ref{table:briar}. PETAL\textit{face} shows significant improvements in performance on BRIAR, with a FAR of $35.12$, $55.35$, and $75.43$ at TAR of $0.01$\%, $0.1$\%, and $1$\%, respectively. It achieves a rank-1 accuracy of $67.42$\%, which is a phenomenal $12.11$\% improvement. The same trend follows for rank-5 and rank-20 accuracies, with improvements of $13.45$\% and $12.44$\%, respectively. Again, we see that full-finetuning does not lead to performance improvements, as discussed in Section~\ref{subsection:tinyface_results}. The large gradient updates in the initial iterations lead to poor convergence. However, PETAL\textit{face} provides significant improvements because of the separate proxy encoders for high-resolution and low-resolution images. It provides meaningful discriminative features for both domains, and the results reiterate the same.

We validate the generalization ability of the proposed PETAL\textit{face} by evaluating it on the IJB-S dataset. We fine-tune the models on the BRIAR train set and test them on IJB-S to gauge the generalization of PETAL\textit{face}. It provides significant improvements in TPIR and rank-retrieval accuracies. It achieves a TPIR of $12.25$\% and $25.28$\% at FPIRs of $1$\% and $10$\%, respectively. The rank-1, rank-5, and rank-10 retrieval accuracies are $38.32$\%, $51.50$\%, and $57.05$\%, respectively. We also see improvements in the \textit{Surveillance-to-Single} and \textit{Surveillance-to-Booking} evaluation settings of the IJB-S dataset, whose results are included in Appendix~\ref{sec:ijbs_results}. One common observation across both datasets is that performance improves as we scale up the pre-training dataset size.

\section{Ablation Studies}
We conduct all ablation studies using a Swin-B model trained on the WebFace4M dataset with CosFace loss. For these experiments, we used static LoRA modules instead of the adaptive LoRA model.
\begin{table}[!tbp]
\centering
\resizebox{0.46\textwidth}{!}{
\begin{tabular}{@{}l ccc cc@{}}
\toprule
\multirow{2}{*}{\textbf{Layers}} & \multicolumn{3}{c}{\textbf{TinyFace}~\cite{cheng2019low}} & \multirow{2}{*}{\begin{tabular}[c]{@{}c@{}}\textbf{Total Model} \\ \textbf{Params}\end{tabular}} & \multirow{2}{*}{\begin{tabular}[c]{@{}c@{}}\textbf{Trainable} \\ \textbf{Params}\end{tabular}} \\
\cmidrule(lr){2-4}
& \textbf{Rank-1} & \textbf{Rank-5} & \textbf{Rank-10} & & \\
\midrule
Pretrained & 72.74 & 76.79 & 79.18 & 213.67M & 213.67M \\
Full Finetuning & 71.32 & 76.42 & 79.45 & 213.67M & 213.67M \\
Attention & 75.59 & 78.83 & 82.13 & 214.23 M & 730k \\
Attention + MLP + Proj + Feature & 74.89 & 78.64 & 81.35 & 216.24M & 2737k \\
Attention + MLP + Proj + Patch Reduction + Feature & 75.16 & 78.64 & 81.59 & 215.96M & 2455k \\
Attention + MLP + Feature & 75.21 & 78.72 & 81.06 & 215.96M & 2455k \\
\rowcolor[gray]{0.93}
\textbf{Attention + Feature} & \textbf{75.64} & \textbf{78.86} & \textbf{81.59} & \textbf{214.54M} & \textbf{1041k} \\
\bottomrule
\end{tabular}}
\caption{Performance of Swin-B models when LoRA is added at different positions in the transformer network.}
\label{table:layer_ablation}
\end{table}
\begin{table}[!tbp]
\centering
\resizebox{0.47\textwidth}{!}{
\begin{tabular}{@{}l ccc cc@{}}
\toprule
\multirow{2}{*}{\textbf{Rank}} & \multicolumn{3}{c}{\textbf{TinyFace}~\cite{cheng2019low}} & \multirow{2}{*}{\begin{tabular}{@{}c@{}}\textbf{Total Model} \\ \textbf{Params}\end{tabular}} & \multirow{2}{*}{\begin{tabular}{@{}c@{}}\textbf{Trainable} \\ \textbf{Params}\end{tabular}} \\
\cmidrule(lr){2-4} 
& \textbf{Rank-1} & \textbf{Rank-5} & \textbf{Rank-10} & & \\
\midrule
2 & 75.61 & 79.02 & 81.73 & 213.88M & 384k \\
4 & 75.56 & 79.05 & 81.59 & 214.10M & 603k \\
\rowcolor[gray]{0.93}
\textbf{8} & \textbf{75.64} & \textbf{78.86} & \textbf{81.59} & \textbf{214.54M} & \textbf{1041k} \\
16 & 75.26 & 79.15 & 81.46 & 215.41M & 1918k \\
32 & 75.45 & 78.94 & 81.81 & 217.17M & 3671k \\
64 & 75.05 & 78.72 & 81.08 & 220.67M & 7177k \\
128 & 75.24 & 78.70 & 81.22 & 227.69M & 14.19M \\
\bottomrule
\end{tabular}}
\caption{Performance of Swin-B models fine-tuned using LoRA modules of varying ranks.}
\label{table:rank_ablation}
\vspace{-10pt}
\end{table}

\textbf{Effect of applying LoRA to different layers in the network:} 
We experimented with adding low-rank decomposition modules at various position within the transformer block.~\cite{hu2021lora} proposed adding LoRA parallel to the attention layers. From our experiments shown in Table~\ref{table:layer_ablation}, we observe that adding LoRA to the final feature layer along with the attention layer leads to superior performance. This adjustment in the final feature layers help align the extracted features based on the updated attention layers, resulting in better overall performance. Moreover, we don't see a drastic increase in the number of trainable parameters, which increased from $730$k to $1041$k, representing only a $0.48$\% increase of total parameters. \textbf{Effect of LoRA rank on performance:} We ablate over different ranks for low-rank decomposition, with the results summarized in Table~\ref{table:rank_ablation}. Our findings indicate that rank-8 yields the best performance, and thus we adopted this rank for training all our models.~\cite{hu2021lora} shows that the attention matrix has an intrinsic low rank, often resulting in better performance with smaller ranks. This is corroborated by the results in Table~\ref{table:rank_ablation}, where ranks $2$, $4$, and $8$ outperform ranks $32$, $64$, and $128$. \textbf{Effect of different backbones on performance:} We conduct experiments with various backbones to demonstrate the broad applicability of PETAL\textit{face}. The results, summarized in Table~\ref{table:iqa}, show that PETAL\textit{face} with the ViT backbone follows a similar trend, outperforming full fine-tuning of the models.
\begin{table}[!tbp]
\centering
\resizebox{0.35\textwidth}{!}{
\begin{tabular}{@{}l ccc@{}}
\toprule
\multirow{2}{*}{\textbf{Training}} & \multicolumn{3}{c}{\textbf{TinyFace}~\cite{cheng2019low}} \\
\cmidrule(lr){2-4}
& \textbf{Rank-1} & \textbf{Rank-5} & \textbf{Rank-10} \\
\midrule
\multicolumn{4}{c}{ViT Backbone with CosFace~\cite{wang2018cosface} Loss Function}\\
Pretrained & 73.57 & 76.95 & 78.94  \\
Full Finetuning & 71.08 & 76.09 & 79.42 \\
LoRA & 73.92 & 77.11 & 79.15\\
PETAL\textit{face} & \textbf{74.14} & \textbf{77.22} & \textbf{79.56} \\
\midrule
\multicolumn{4}{c}{Ablation using different IQA networks}\\
Pretrained & 72.74 & 76.79 & 79.18  \\
Full Finetuning & 71.32 & 76.42 & 79.45 \\
PETAL\textit{face} (BRISQUE)~\cite{mittal2012no} & 75.16 & 78.46 & 80.90 \\
PETAL\textit{face} (CR-FIQA)~\cite{boutros2023cr} & 75.34 & 78.75 & 81.30 \\
\rowcolor[gray]{0.93}
PETAL\textit{face} (CNN-IQA)~\cite{kang2014convolutional} & \textbf{75.64} & \textbf{78.86} & \textbf{81.59} \\
\bottomrule
\end{tabular}}
\caption{Results using ViT Backbone and PETAL\textit{face} performance using different image quality estimators.}
\label{table:iqa}
\vspace{-8pt}
\end{table}
\textbf{Effect of Image quality assessment on performance:}
We experimented with two lightweight NR-IQA models: BRISQUE~\cite{mittal2012no} and CNN-IQA~\cite{kang2014convolutional}, and one face image quality assessment network CR-FIQA~\cite{boutros2023cr}. We present the corresponding results in Table \ref{table:iqa}. The performance on TinyFace when using BRISQUE as the IQA yielded rank-1, rank-5, and rank-10 retrieval accuracies of 75.16\%, 78.46\%, and 80.90\%, respectively. While CR-FIQA~\cite{boutros2023cr} showed competitive results, its performance was slightly lower than CNN-IQA, likely because it is trained on MS1MV2~\cite{deng2019arcface} that doesn't contain diverse range of degradation that are present in challenging evaluation datasets like TinyFace, BRIAR and IJB-S. CNN-IQA demonstrated superior performance, leading us to select CNN-IQA as our IQA for all subsequent experiments. We emphasize that this boost in performance is due to the robustness of a CNN-IQA method arising due to its training.
\section{Limitation and Future work}
As described in section \ref{sec:proposed}, PETAL\textit{face} utilizes an off the shelf IQA module to model the parameter $\alpha$ defining the strength of the chosen LoRA module. However, most image quality assessment networks are not accurate. Hence, research on better IQA models would enable a boost in the performance of PETAL\textit{face}. Moreover, we define a manually selected heuristic for the choice of parameter $\alpha$ for choosing the LoRA module. However, one may perform a more sophisticated heuristic selection by a parameter sweep over a validation set. We leave these challenges as open problems to be addressed in future work. 

\section{Conclusion}
In this paper, we propose PETAL\textit{face}, a new method that harnesses the power of parameter-efficient fine-tuning to address the challenging problem of low-resolution face recognition. To achieve this, we introduce a novel image quality assessment based twin LORA module, which significantly enhances the model's ability to handle varying image qualities. By adopting this design choice, we effectively tackle two major issues prevalent in existing works: catastrophic forgetting and the domain difference between gallery and probe images. We conduct extensive experiments across multiple benchmarks on low-resolution datasets and achieve state-of-the-art results across various metrics. Notably, our approach also preserves performance on high-resolution and mixed-quality datasets. Models fine-tuned using PETAL\textit{face} demonstrates versatility and can serve as a generalized model capable of handling a wide range of image resolutions, making it highly suitable for real-world deployment and practical applications. 

\section{Acknowledgement}
This research is based upon work supported in part by the Office of the Director
of National Intelligence (ODNI), Intelligence Advanced Research Projects Ac-
tivity (IARPA), via [2022-21102100005]. The views and conclusions contained
herein are those of the authors and should not be interpreted as necessarily
representing the official policies, either expressed or implied, of ODNI, IARPA,
or the U.S. Government. The US Government is authorized to reproduce and
distribute reprints for governmental purposes notwithstanding any copyright an-
notation therein.

{\small
\bibliographystyle{ieee_fullname}
\bibliography{egbib}
}

\newpage
\onecolumn
\thispagestyle{empty}
\appendix

\begin{center}
    {\Large \textbf{Appendix}}
\end{center}

\noindent The Appendix is organized into the following sections. First, we discuss additional implementation details. Next, we present the results on other evaluation settings of the IJB-S dataset. Also, we present a gradient analysis of PETAL\textit{face} and compare it to full fine-tuning to highlight that the proposed approach leads to stable convergence. Finally, we provide a failure case analysis of PETAL\textit{face}.
\section{Implementation Details}
\label{sec:implementation}
All deployment codes were implemented in PyTorch framework and executed it on eight A5000 GPUs, each equipped with 24GB of memory.The models are trained using the AdamW optimizer and a polynomial learning rate (LR) scheduler, with an initial learning rate of $5e^{-4}$ and a weight decay set to $0.1$. We fine-tuned for 40 epochs on TinyFace~\cite{cheng2019low} dataset , utilizing a warm-up of 2 epochs and a batch size of 8. For the BRIAR~\cite{cornett2023expanding} dataset, we fine-tuned for 10 epochs with one warm-up epoch, also using a batch size of $8$. We utilized a low-rank decomposition of 8 for the TinyFace dataset and $32$ for the BRIAR dataset. We employed CNN-IQA~\cite{kang2014convolutional} as our NR-IQA network to assign weightages to the LoRA modules. We present the implementation code modules for the adaptive weight estimated and Adaptive LoRA in the below code fragments. The weightage for the twin LoRA modules is calculated using \textcolor{blue}{\code{generate\_alpha}}. The final output is calculated as shown in \textcolor{blue}{\code{adaptive\_lora}}. 
The complete PETAL\textit{face} training framwork for a single layer is outlined in Algorithm~\ref{alg:petalface}.

\begin{algorithm}
\caption{PETAL\textit{face} Training Framework for a single layer}
\label{alg:petalface}
\begin{algorithmic}[1]
\STATE \textbf{Given:} Pre-trained weight matrix \( W_0 \in \mathbb{R}^{m \times n} \), LoRA blocks \( W_1 \in \mathbb{R}^{m \times n} \) and \( W_2 \in \mathbb{R}^{m \times n} \), Input images \( X = \{ x_i \mid 0 \leq i < p \} \), Image quality estimator \(\phi(x)\)

\FOR{each dataset}
    \STATE Sample a random \( l \) number of samples \( x_1, x_2, \ldots, x_l \)
    \STATE Calculate the mean \( \mu \) and standard deviation \( \sigma \) of the quality scores:
    \[
    \mu = \frac{1}{l} \sum_{i=1}^{l} \phi(x_i), \quad
    \sigma = \sqrt{\frac{1}{l} \sum_{i=1}^{l} (\phi(x_i) - \mu)^2}
    \]
\ENDFOR
    \STATE Set the threshold \( t = \mu + \sigma \)
    \FOR{each sample \( x_i \) in X}
        \STATE $q_i = \phi(x_i)$ 
        \STATE The weightage $\alpha_i$ is calculated using $q_i$ by:
        \[
        \alpha_i = 
        \begin{cases} 
        0.5 & \text{if } q_i = t \\
        0.5 - (t - q_i) & \text{if } q_i < t \\
        0.5 + (q_i - t) & \text{if } q_i > t 
        \end{cases}
        \]
    \ENDFOR
\STATE We obtain image quality scores \( Q = \{ q_i \mid 0 \leq i < p \ni q_i = \phi(x_i), \forall \ 0 \leq i < p \} \)
\STATE The final output $x_{out}^{i}$ is calculated as:
\[
x_{out}^i = W_0(x_i) + \alpha_i W_1(x_i) + (1 - \alpha_i) W_2(x_i)
\]
\end{algorithmic}
\end{algorithm}

\newpage
\begin{center}
\vspace{0.5em}
\captionsetup{type=listing}
\caption*{Image quality based weight assignment}
\begin{small}
\begin{lstlisting}[language=Python, mathescape]
!pip install pyiqa
iqa = pyiqa.create_metric('cnniqa').cuda()

def generate_alpha(img, iqa, threshold):
    device = img.device
    BS, C, H, W = img.shape
    alpha = torch.zeros((BS, 1), dtype=torch.float32, device=device)

    score = iqa(img)
    for i in range(BS):
        if score[i] == threshold:
            alpha[i] = 0.5 
        elif score[i] < threshold:
            alpha[i] = 0.5 - (threshold - score[i])
        else:
            alpha[i] = 0.5 + (score[i] - threshold)
    return alpha
\end{lstlisting}
\end{small}
\vspace{0.5em}
\end{center}

\begin{center}
\captionsetup{type=listing}
\caption*{Adaptive LoRA}
\begin{small}
\begin{lstlisting}[language=Python, mathescape]
class AdaptiveLoRA(nn.Linear):
    def __init__(self, in_features: int, out_features: int,  r: int, scale: int, bias: bool=True) -> None:
        super().__init__(in_features, out_features, bias)
        # LoRA 1
        self.r_1 = r
        self.scale_1 = scale
        self.trainable_lora_down_1 = nn.Linear(in_features, self.r_1, bias=False)
        self.dropout_1 = nn.Dropout(0.1)
        self.trainable_lora_up_1 = nn.Linear(self.r_1, out_features, bias=False)
        self.selector_1 = nn.Identity()
        nn.init.normal_(self.trainable_lora_down_1.weight, std=1/self.r_1)
        nn.init.zeros_(self.trainable_lora_up_1.weight)

        # LoRA 2
        self.r_2 = r
        self.trainable_lora_down_2 = nn.Linear(in_features, self.r_2, bias=False)
        self.dropout_2 = nn.Dropout(0.1)
        self.trainable_lora_up_2 = nn.Linear(self.r_2, out_features, bias=False)
        self.scale_2 = scale
        self.selector_2 = nn.Identity()

        nn.init.normal_(self.trainable_lora_down_2.weight, std=1/self.r_2)
        nn.init.zeros_(self.trainable_lora_up_2.weight)

    def forward(self, x, alpha):
        out = F.linear(x, self.weight, self.bias)
        lora_adjustment_1 = self.scale_1*self.dropout_1(self.trainable_lora_up_1( self.selector_1(self.trainable_lora_down_1(x))))
        lora_adjustment_2 = self.scale_2*self.dropout_2(self.trainable_lora_up_2( self.selector_2(self.trainable_lora_down_2(x))))
        out = out + (1 - alpha)*lora_adjustment_1 + alpha*lora_adjustment_2
        return  out
\end{lstlisting}
\end{small}
\vspace{0.5em}
\end{center}

\section{IJB-S Results}
\label{sec:ijbs_results}

\begin{table*}[h]
\begin{center}
\resizebox{\textwidth}{!}{
\begin{tabular}{@{}lcc ccc ccc@{}}
\toprule[0.15em]
\multirow{2}{*}{\textbf{Training}}  & \multirow{2}{*}{\textbf{Dataset}}& \multirow{2}{*}{\textbf{Arch.}} & \multicolumn{3}{c}{\textbf{IJB-S (Surveillance to Single)~\cite{kalka2018ijb}}} & \multicolumn{3}{c}{\textbf{IJB-S (Surveillance to Booking)~\cite{kalka2018ijb}}} \\
\cmidrule(lr){4-6} \cmidrule(lr){7-9}
& & & Rank-1 & Rank-5 & Rank-10 & Rank-1 & Rank-5 & Rank-10\\
\midrule[0.15em] 
Pre-trained & WBF4M~\cite{zhu2021webface260m} & R50 & 32.01 & 45.72 & 51.25 & 43.82 & 55.75 & 61.28 \\
Pre-trained & WBF4M~\cite{zhu2021webface260m} & Swin-B & 33.23 & 49.85 & 57.63 & 46.22 & 59.40 & 64.93 \\
\midrule[0.15em]
Full-FT & WBF4M~\cite{zhu2021webface260m} & Swin-B  & 4.20 & 10.95 & 16.64 & 5.39 & 13.31 & 19.84 \\
\rowcolor[gray]{0.93}
\textbf{PETAL\textit{face}} & WBF4M~\cite{zhu2021webface260m} & Swin-B & \textcolor{blue}{\textbf{37.12}} & \textcolor{blue}{\textbf{51.07}} & \textcolor{blue}{\textbf{57.60}} & \textcolor{blue}{\textbf{43.63}} & \textcolor{blue}{\textbf{59.85}} & \textcolor{blue}{\textbf{66.15}} \\
\midrule
\rowcolor[gray]{0.93}
\textbf{PETAL\textit{face}} & WBF12M~\cite{zhu2021webface260m} & Swin-B  & 44.40 & 57.84 & 63.87 & 51.09 & 64.67 & 70.30 \\
\bottomrule[0.15em]
\end{tabular}
}

\caption{Results on IJB-S~\cite{kalka2018ijb} dataset in \textit{Surveillance-to-single} and \textit{Surveillance-to-booking} settings. The models are fine-tuned on the BRIAR train set. We report the closed-set rank retrieval (Rank-1, Rank-5 and Rank-10). [\textcolor{blue}{\textbf{BLUE}}] indicates the best results for models trained on WebFace4M~\cite{zhu2021webface260m}.}
\label{table:ijbs}
\end{center}

\end{table*}

The results on the IJB-S dataset in the \textit{Surveillance-to-single} and \textit{Surveillance-to-booking} settings are shown in Table~\ref{table:ijbs}. In the \textit{Surveillance-to-single} setting, gallery images are single high-quality images. Similarly, in \textit{Surveillance-to-booking}, we have high-quality gallery images from different angles. The probes are of surveillance quality in both settings. This setup highlights the importance of having two proxy encoders for different resolutions within the same backbone, which are weighted based on input image quality. PETAL\textit{face} shows improved performance with rank-1, rank-5, and rank-10 retrieval accuracies of $44.40$, $57.84$, and $63.87$, respectively, in the \textit{Surveillance-to-single} setting. We see similar improvements in the \textit{Surveillance-to-booking} setting for rank-5 and rank-10 accuracies, with increases of $0.45$\% and $1.22$\%. The results demonstrate the generalization capability of the proposed fine-tuning approach. Although the model is fine-tuned on the BRIAR dataset, the knowledge of low-resolution data gained from that can be translated to other datasets such as IJB-S. Additionally, we observe a significant drop in performance when we fully fine-tune the model. 
We discussed the causes in the main paper and want to reiterate here. Face recognition models are pre-trained on large datasets with high-resolution images. When fine-tuning on low-resolution datasets, the model encounters a domain difference, which leads to large gradient updates initially. This deviates the model from the original pre-trained state abruptly, leading to poor convergence. We provide a gradient analysis in Section~\ref{sec:grad} to validate our claims.

\section{Gradient Analysis}
\label{sec:grad}

\begin{figure}[t]
    \centering
    \begin{subfigure}[b]{0.49\textwidth}
        \centering
        \includegraphics[width=\textwidth]{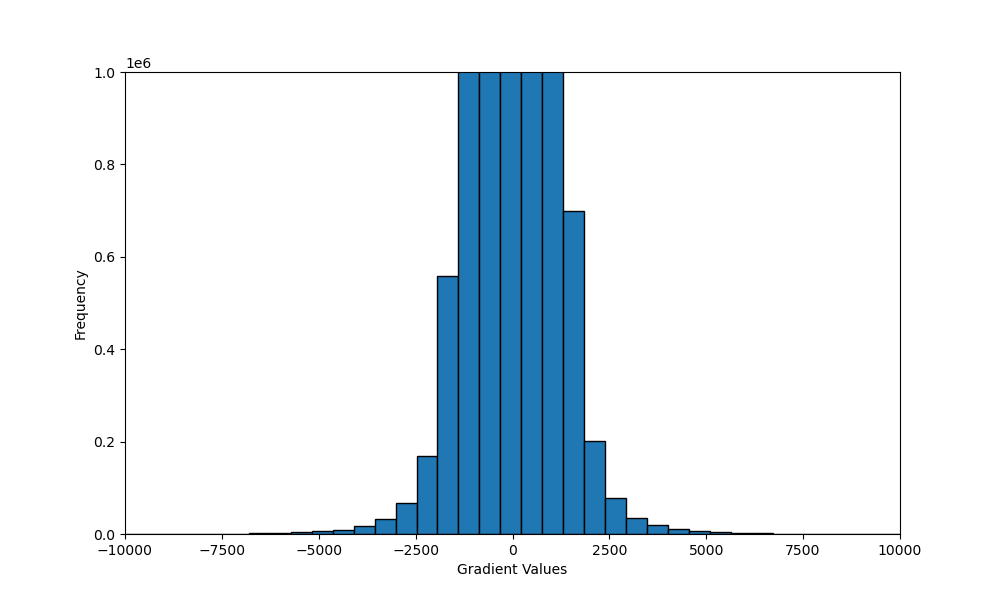}
        \caption{Full fine-tuning}
        \label{fig:image1}
    \end{subfigure}
    \hfill
    \begin{subfigure}[b]{0.49\textwidth}
        \centering
        \includegraphics[width=\textwidth]{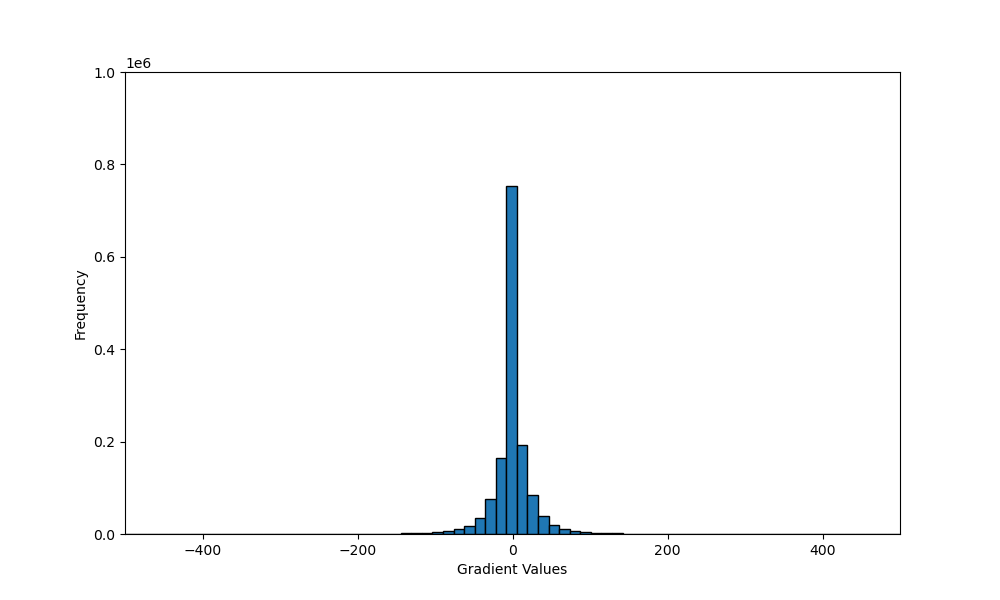}
        \caption{PETAL\textit{face}}
        \label{fig:image2}
    \end{subfigure}
    \caption{Comparison of initial gradients when (a) Full fine-tuning a model and using (b) PETAL\textit{face} fine-tuning approach. We can see that PETAL\textit{face} has small initial gradients which results in stable and gradual convergence. \textbf{NOTE:} The scale of the 'Gradient Values' axis for Full fine-tuning and PETAL\textit{face} is different.}
    \label{fig:grad}
\end{figure}

We analyze the gradients of the model backbone when fully fine-tuning the model versus when using PETAL\textit{face} to fine-tune the model. We plot the frequency of gradient values for the first iteration of training. As shown in the Figure~\ref{fig:grad}, we see that when fully fine-tuning the model, the initial gradients are very large, and even after clipping the gradients, there will be a large number of parameters that will change significantly. This is due to the domain difference between pre-trained and fine-tuned data, leading to an abrupt deviation from pre-trained weights when fully fine-tuning the model. The initial value of gradients when using the PETAL\textit{face} fine-tuning approach results in relatively smaller gradients initially, leading to more stable and gradual convergence and improved performance. Moreover, the original weights remain frozen thereby preserving all information learned during large scale training. This demonstrates the superiority of our approach in efficiently adapting to low-resolution data.

\section{Failure Case Analysis}
\begin{figure*}[!h]
    \centering
    \includegraphics[width=0.9\linewidth]{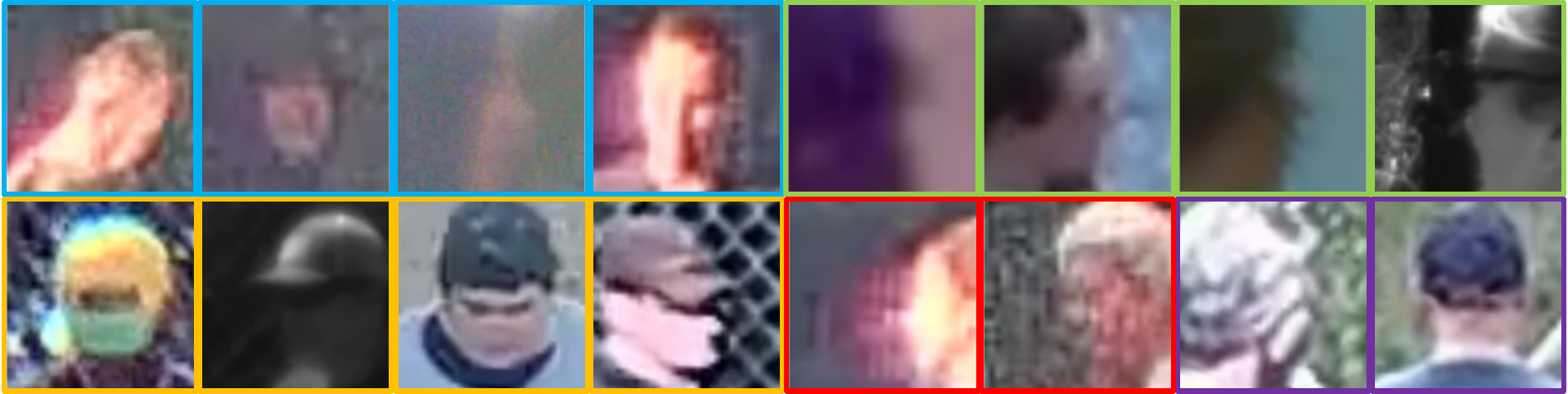}
    \caption{Failure Case Analysis of PETAL\textit{face} on the BRIAR dataset. All the subjects are consented for publication.}
    \label{fig:failure}
\end{figure*}

We conducted a failure case analysis of the probe videos, as summarized in Figure~\ref{fig:failure}, to examine the limitations of our model. We found that it struggled to recognize faces that were  \textcolor{ProcessBlue}{very low in resolution} and featured \textcolor{LimeGreen}{extreme head poses}. It also failed in cases of \textcolor{BurntOrange}{heavy occlusion}, where faces were obscured by items like caps, masks, or sunglasses. Additionally, the model performed poorly when faces were degraded by \textcolor{red}{atmospheric turbulence}, making recognition difficult. Furthermore, the model failed with probe videos \textcolor{Fuchsia}{lacking frontal face} views, as it could not identify individuals without clear frontal visibility throughout the video.

\end{document}